\documentclass[conference]{IEEEtran}
\IEEEoverridecommandlockouts

\usepackage{cite}
\usepackage{amsmath,amssymb,amsfonts}
\usepackage{graphicx}
\usepackage{textcomp}
\usepackage{xcolor}
\usepackage{booktabs}
\usepackage{multirow}
\usepackage{tabularx}
\usepackage{url}

\def\BibTeX{{\rm B\kern-.05em{\sc i\kern-.025em b}\kern-.08em
    T\kern-.1667em\lower.7ex\hbox{E}\kern-.125emX}}

\begin{document}
\title{Beyond Recall Saturation: Scaling Hard-Negative Reranker Training for Industrial Legal Retrieval}
\author{
\IEEEauthorblockN{Yunhan Li}
\IEEEauthorblockA{\textit{SIAT-DELI AI and Law Joint Lab,} \\ 
Shenzhen Institutes of Advanced Technology, \\ 
Chinese Academy of Sciences\\
Shenzhen, China\\
\textit{City University of Macau}\\
Macau, China\\
D24092110205@cityu.edu.mo}
\and
\IEEEauthorblockN{Mingjie Xie}
\IEEEauthorblockA{\textit{SIAT-DELI AI and Law Joint Lab,} \\ 
Shenzhen Institutes of Advanced Technology, \\ 
Chinese Academy of Sciences \\
Shenzhen, China\\
suat25060313@stu.suat-sz.edu.cn}
\and
\IEEEauthorblockN{Zeyang Shi}
\IEEEauthorblockA{\textit{City University of Macauy}\\
Shenzhen, China}
\and
\IEEEauthorblockN{Gengshen Wu\IEEEauthorrefmark{1}}
\IEEEauthorblockA{\textit{City University of Macau}\\
Macau, China\\
gswu@cityu.edu.mo}
\and
\IEEEauthorblockN{Min Yang\IEEEauthorrefmark{1}}
\IEEEauthorblockA{\textit{SIAT-DELI AI and Law Joint Lab,} \\ 
Shenzhen Institutes of Advanced Technology, \\ 
Chinese Academy of Sciences \\
Shenzhen, China\\
min.yang@siat.ac.cn}
\thanks{\IEEEauthorrefmark{1}Corresponding authors.}
}
\maketitle
\begin{abstract}
Industrial retrieval systems often improve quality by enlarging embedding candidate pools or adopting complex training pipelines. We show that, under a sufficiently large but practical candidate pool, embedding recall exhibits diminishing returns, motivating a closer examination of reranker quality. On Chinese legal retrieval benchmarks, embedding top-300 reaches a high-recall operating region, while relevant evidence may still be ranked below the output cutoff. We compare hard-negative strategies for reranker training, including embedding-mined, self-mined, and dynamic-negative variants, under aligned retrieval protocols that evaluate reranking after candidate retrieval. Among the evaluated staged configurations, an embedding-to-self-mined pipeline provides a favorable trade-off; additional dynamic stages do not improve all held-out metrics. The selected pipeline scales from a 0.6B to a 4B reranker. Relative to the native 4B reranker, the trained 4B Stage2 model improves reported metrics on the STARD and CSAID benchmarks and on an internal development set, and is preferred in 72.4\% of non-tied comparisons on 410 de-identified, reconstructed internal operational-query inputs in a latency-sensitive offline preference evaluation. These results support evaluating the retrieval and reranking stages separately when selecting an optimization target for industrial retrieval systems.
\end{abstract}

\section{Introduction}

Legal AI products increasingly depend on retrieval systems as shared infrastructure. A single retrieval service may support statute search, legal question answering, and retrieval-augmented generation (RAG)-based analysis. In these settings, retrieval quality directly affects user-visible evidence quality, while retrieval latency and model maintenance cost constrain deployment choices. The common architecture is a two-stage pipeline: a dense embedding model recalls a candidate pool, and a reranker reorders the candidates before returning a compact evidence set.

This architecture creates a practical optimization question: when retrieval quality is unsatisfactory, should effort go into increasing recall through larger candidate pools, or improving the reranker? The answer matters in production because larger candidate pools increase reranking cost, while increasingly complicated training pipelines can be difficult to reproduce, monitor, and scale. The problem is especially salient in legal retrieval, where queries are often informal but relevant evidence depends on fine-grained legal distinctions.

\begin{figure*}[t]
    \centering
    \includegraphics[width=\linewidth,height=0.30\textheight,keepaspectratio]{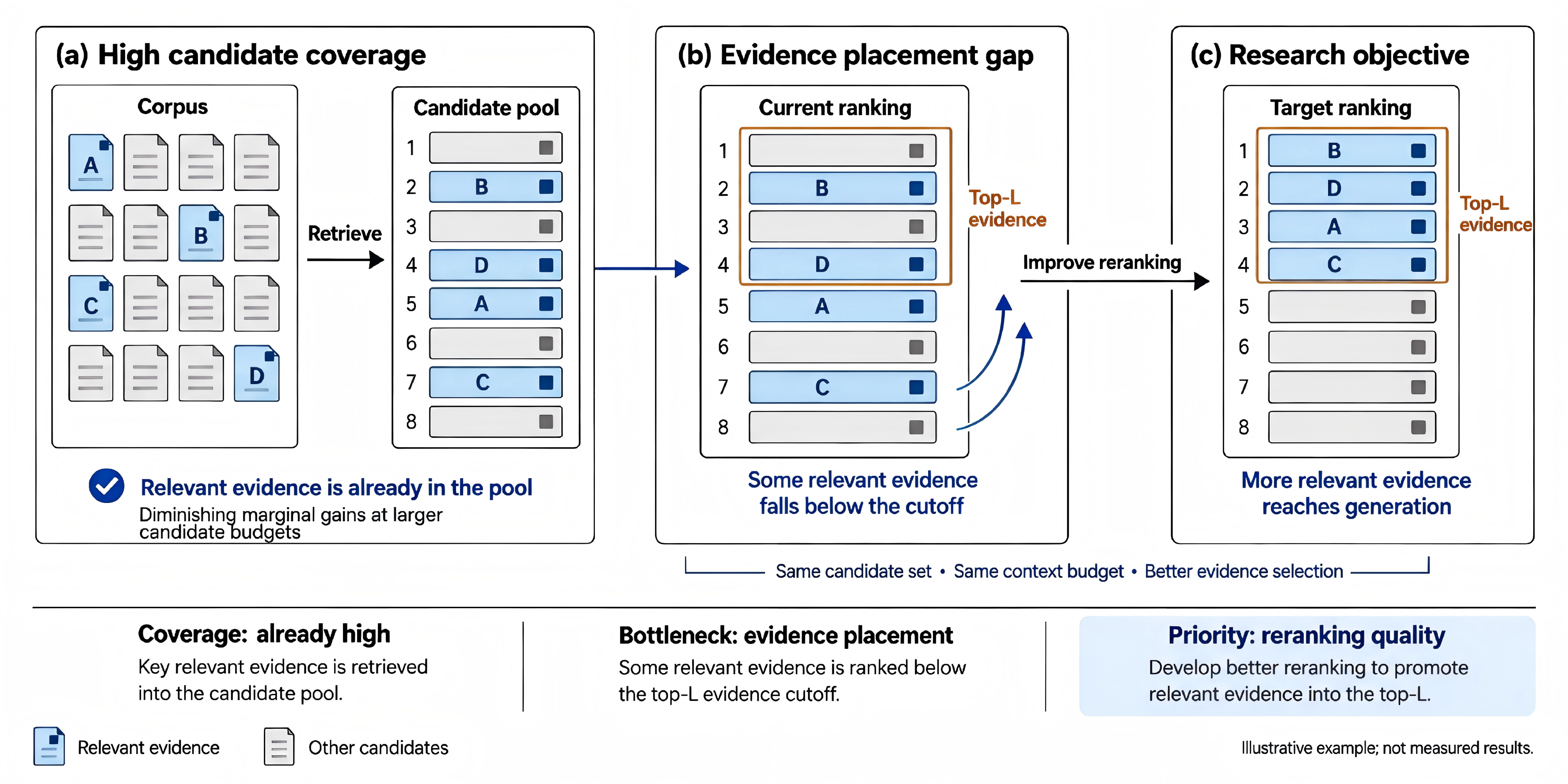}
    \caption{From candidate coverage to evidence placement. (a) Retrieval places relevant evidence in a candidate pool. (b) Some retrieved evidence remains below the top-$L$ cutoff. (c) The research objective is to improve reranking so that more relevant evidence enters the final list under the same candidate set and output budget. Blue documents denote relevant evidence; gray documents denote other candidates. The rankings and complete coverage shown in this example are illustrative, not measured results or a claim of perfect retrieval or reranking.}
    \label{fig:coverage_to_reranking}
\end{figure*}

Figure~\ref{fig:coverage_to_reranking} illustrates the distinction between candidate coverage and evidence placement that motivates our study. When retrieval already places a large proportion of relevant evidence in the candidate pool, some of that evidence may still fall below the top-$L$ cutoff used by search or downstream generation. Improving its placement can therefore increase the relevant evidence delivered within a fixed output budget, even without changing the candidate set. This motivates investigating reranker supervision in a high-coverage setting, while recognizing that reranking cannot recover documents missing from the pool. Section~\ref{sec:recall} examines the candidate-coverage premise quantitatively.

In this paper, we study this question in an industrial Chinese legal retrieval system. Rather than assuming that larger candidate pools or increasingly aggressive reranker adaptation strategies always improve downstream retrieval quality, we evaluate retrieval optimization component by component. We first analyze embedding recall under candidate pools up to top-300 and find that recall gains become increasingly marginal at larger retrieval budgets. We then focus subsequent optimization on reranker adaptation and compare multiple hard-negative construction strategies, including embedding-mined negatives, self-mined  negatives, and dynamic refresh variants. Finally, we test whether the selected reranker training recipe scales from 0.6B to 4B and whether the resulting gains transfer to de-identified, reconstructed operational-query inputs under an offline preference-evaluation setting.

Our contributions are:
\begin{itemize}
\item We provide an industrial analysis showing diminishing embedding-recall gains under a sufficiently large candidate pool on legal retrieval benchmarks, motivating reranker adaptation as a complementary optimization target.

\item We compare multiple hard-negative construction strategies and show that, among the evaluated staged configurations, embedding-to-reranker hard-negative adaptation provides a favorable trade-off, whereas increasingly dynamic refresh variants do not improve all held-out metrics.

\item We demonstrate that the selected reranker-focused training pipeline scales consistently from a 0.6B reranker to a stronger 4B reranker, yielding consistent improvements over the native 4B reranker across held-out evaluation sets.

\item We complement benchmark metrics with a randomized LLM preference evaluation on 410 de-identified, reconstructed internal operational-query inputs, where the trained 4B reranker is preferred in 72.4\% of non-tied comparisons under a latency-sensitive offline evaluation setting.
\end{itemize}
Code and implementation details are publicly available at \url{https://github.com/lyxx3rd/beyond-recall-saturation}.

\section{Related Work}

\subsection{Dense Retrieval and Reranking}

Dense retrieval has become a standard foundation for neural search systems, where embedding models provide efficient candidate generation and rerankers improve final ranking quality~\cite{zhao2024dense,zhang2025qwen3}. Retrieval-augmented generation (RAG) systems further increase the importance of retrieval quality because downstream generation depends on the retrieved evidence~\cite{lewis2020retrieval}. Our work follows this two-stage architecture but focuses on how optimization effort should be allocated once retrieval recall becomes sufficiently high.

\subsection{Hard-Negative Mining and Adaptive Supervision}

Hard negatives are widely used to improve contrastive retrieval and ranking models because they expose the model to increasingly confusable candidates~\cite{nian2025w,hao2025domain}. Prior work also explores curriculum learning and progressively harder supervision strategies~\cite{bengio2009curriculum}. Multi-stage fine-tuning has also been explored in retrieval and continual adaptation settings, where earlier stages often rely on embedding-based hard negatives before later-stage reranker refinement~\cite{ma2024fine,guan2025multi}. Recent PEFT and continual-learning studies further investigate sequential LoRA adaptation and staged parameter-efficient updates across tasks or training phases~\cite{jin2025multi}. 

More recent approaches further investigate iterative or dynamically refreshed negative mining to improve retrieval adaptation~\cite{zhan2021optimizing, meghwanietal2025hard}. Our work differs in focusing on the stability and scalability of adaptive hard-negative refresh strategies under industrial reranking pipelines.

\subsection{Retrieval Optimization under Large Candidate Pools}

Larger candidate pools generally improve retrieval recall but also increase reranking latency and serving cost~\cite{zhao2024dense}. Existing retrieval systems often optimize embedding recall and reranking quality separately, while more recent retrieval optimization methods increasingly rely on hard-negative adaptation and dynamic ranking supervision~\cite{zhan2021optimizing}. However, less attention has been paid to how optimization priorities should be assessed once recall gains flatten at larger candidate pools. Our work studies this trade-off in an industrial legal retrieval setting by examining reranker adaptation under a fixed high-recall candidate budget.

\subsection{Legal Information Retrieval}

Legal retrieval differs from open-domain search because queries may be informal while relevance depends on statutory language and fine-grained legal concepts~\cite{bajvcic2017new,maxwell2008concept,su2024stard}. Recent legal retrieval benchmarks and systems highlight the need for domain-specific evaluation~\cite{li2026legalmalrmultiagentqueryunderstandingllmbased}. Our study contributes deployment-oriented evidence from Chinese legal retrieval workloads.

\section{Retrieval Setting and Motivation}

Our production setting follows a two-stage retrieval design. Given a legal query, an embedding model retrieves a candidate pool from the legal corpus. A generative reranker then scores each query-document pair and returns the top-ranked evidence passages for downstream search or RAG applications. This design separates coverage-oriented recall from fine-grained relevance ranking.

We focus on legal retrieval for two reasons. First, industrial legal queries often use informal descriptions that do not directly match statutory language. Second, relevance may depend on subtle distinctions between legal concepts, making reranking important even when the relevant document is already present in the candidate pool. Our goal is not to claim universal cross-domain generalization, but to provide an industrial case study of retrieval optimization under a domain-specific two-stage architecture.

\subsection{Data Sources}

The training pool contains positive query-evidence pairs from two sources: internally generated or curated legal training data, and the training portion of STARD~\cite{su2024stard}. After cleaning, the pool contains 10,455 query-level samples, with 9,221 from internal legal data and 1,234 from STARD training data. Each query may have multiple positive evidence passages, with an average of 1.90 positive passages per query. We split this cleaned pool into 9,932 training samples and 523 internal validation samples using a fixed 95/5 random split. The internal validation set is used only for training-time model selection and early stopping.

Final benchmark evaluation uses the STARD test set (309 queries) and CSAID (118 queries). EVAL is an internal development set used for configuration comparison during development and is reported separately; it is not treated as a final benchmark. We additionally use 410 de-identified, reconstructed internal operational-query inputs for supplementary preference-level analysis. Table~\ref{tab:data_summary} summarizes the roles of the training, validation, benchmark, and operational-query evaluation data used in this work.

Before negative mining, we remove exact query overlaps between the training pool and the STARD and CSAID benchmark queries. We do not claim document-level, case-level, or temporal isolation; instead, we report the final benchmark and operational-query evaluations separately from the internal validation split.

\begin{table*}[t]
\caption{Overview of the data roles used in this work. Internal validation is used only for training-time checkpoint selection; EVAL is a development set rather than a final benchmark.}
\centering
\scriptsize
\resizebox{\textwidth}{!}{%
\begin{tabular}{llrl}
\toprule
Data & Type & Size & Role \\
\midrule
Internal legal data & Generated/curated query-evidence pairs & 9,221 & Training pool \\
STARD train & Public legal retrieval training data & 1,234 & Training pool \\
STARD test & Public held-out benchmark & 309 & Final benchmark evaluation \\
CSAID & Public held-out benchmark & 118 & Final benchmark evaluation \\
EVAL & Internal development set & -- & Configuration comparison; not a final benchmark \\
Internal operational-query inputs & De-identified and reconstructed & 410 & Preference evaluation \\
\bottomrule
\end{tabular}
}
\label{tab:data_summary}
\end{table*}

\subsection{Embedding Recall Saturation}\label{sec:recall}

\begin{figure*}[t]
  \centering
  \begin{minipage}{0.48\linewidth}
    \centering
    \includegraphics[width=\linewidth]{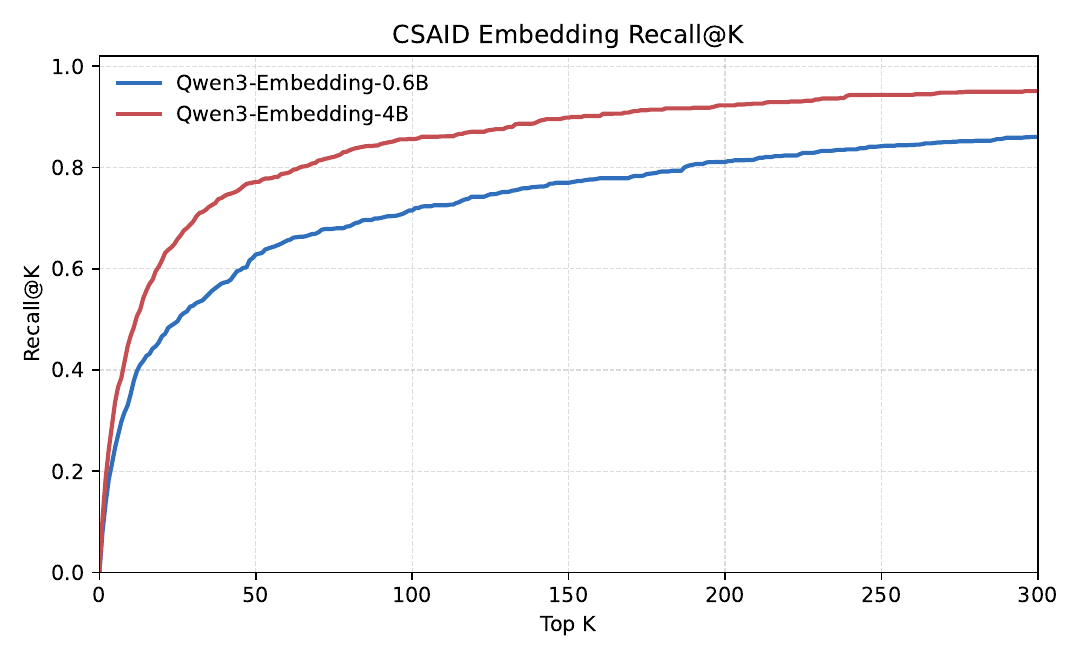}
  \end{minipage}
  \hfill
  \begin{minipage}{0.48\linewidth}
    \centering
    \includegraphics[width=\linewidth]{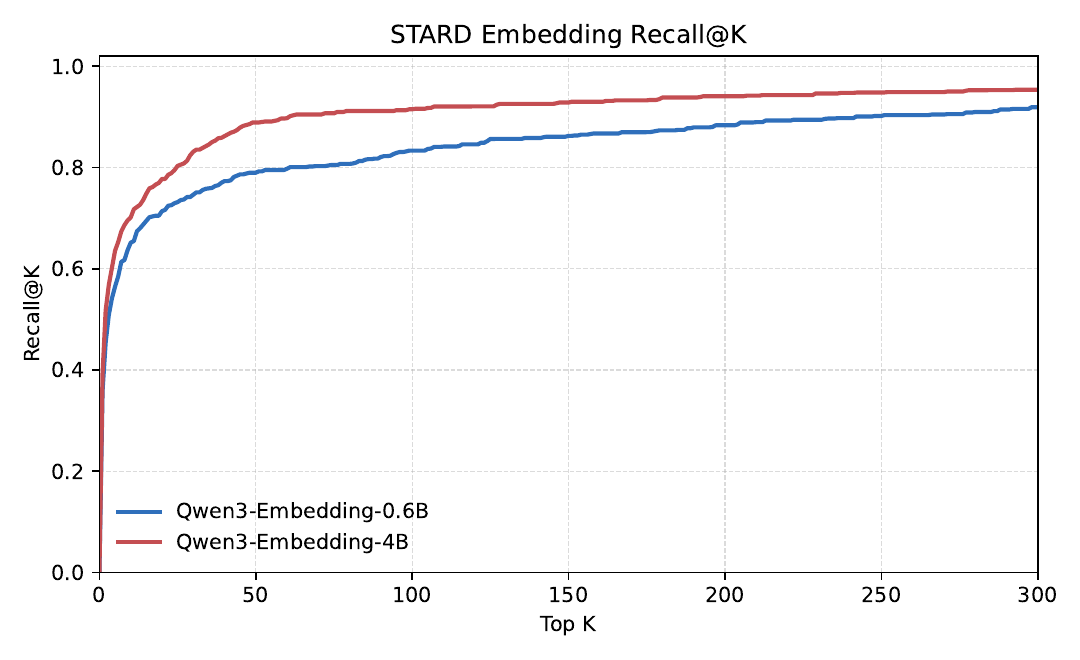}
  \end{minipage}
  \caption{Embedding recall on CSAID and STARD as the candidate pool increases. The curves motivate using a sufficiently large candidate pool for reranker evaluation rather than treating recall expansion as the only optimization target.}
  \label{fig:embedding_recall}
\end{figure*}

Optimizing a two-stage retrieval system requires distinguishing candidate coverage from final evidence ordering. Increasing the embedding candidate pool can recover additional relevant documents, but it also increases the number of query--document pairs that the reranker must score. Before investigating reranker training, we therefore examine how much relevant evidence is already available at different candidate budgets. We compare Qwen3-Embedding-0.6B and Qwen3-Embedding-4B on the STARD and CSAID test sets, with candidate pool sizes from 1 to 300. Figure~\ref{fig:embedding_recall} shows the recall curves, and Table~\ref{tab:embedding_recall_summary} reports selected operating points.

\begin{table*}[t]
\caption{Embedding recall at selected candidate budgets on the test sets. Values are percentages; $\Delta_{250\rightarrow300}$ is the increase in recall from top-250 to top-300, in percentage points.}
\label{tab:embedding_recall_summary}
\centering
\small
\begin{tabular}{llrrrrr}
\toprule
Embedding model & Dataset & R@100 & R@200 & R@250 & R@300 & $\Delta_{250\rightarrow300}$ \\
\midrule
Qwen3-0.6B & STARD & 83.31 & 88.31 & 90.17 & 91.89 & 1.73 \\
Qwen3-0.6B & CSAID & 71.48 & 81.07 & 84.26 & 86.03 & 1.77 \\
Qwen3-4B & STARD & 91.47 & 94.07 & 94.81 & 95.34 & 0.54 \\
Qwen3-4B & CSAID & 85.62 & 92.24 & 94.33 & 95.07 & 0.74 \\
\bottomrule
\end{tabular}
\end{table*}

At top-300, the 4B embedding model reaches 95.34\% recall on STARD and 95.07\% on CSAID, indicating that the candidate pool covers a large proportion of the relevant evidence. The 0.6B model reaches 91.89\% and 86.03\%, respectively, so the remaining coverage gap depends on both model capacity and dataset. Recall continues to improve as the pool grows, but the curves flatten toward the upper end of the measured range. For the 4B model, expanding the pool from 250 to 300 candidates adds 50 documents to score while increasing recall by only 0.54 percentage points on STARD and 0.74 on CSAID. These observations indicate diminishing marginal recall gains within the tested range; they do not establish that larger pools beyond 300 would provide no further benefit.

High candidate recall creates an opportunity for reranking, because evidence that is present in the candidate pool may still be ranked too low to appear in the compact list returned to users or downstream RAG components. Candidate coverage and the quality of that final list are therefore distinct: the reranker must promote relevant documents while suppressing plausible but non-relevant alternatives. It cannot recover documents absent from the candidate pool, and the recall values above should not be interpreted as complete coverage for every query or every embedding model. We use top-300 as a controlled candidate budget for studying reranker adaptation, reducing the influence of missing candidates while retaining the remaining coverage limitation. This is an evaluation setting rather than a claim that top-300 is the optimal deployment budget.

This motivates the focus on negative supervision in the following section. The embedding candidate pool contains both relevant evidence and competing documents that the reranker must distinguish. Embedding rankings provide one way to identify difficult non-positive candidates; a fixed or evolving reranker provides alternative judgments of which candidates remain confusing. Our subsequent experiments examine which of these signals yields more effective training supervision and whether their sequential combination improves final evidence ordering.

\section{Method}

We use Qwen3 embedding and reranking models as the base model family~\cite{zhang2025qwen3}. The key design choice is how to construct negatives for reranker training. Legal relevance is fine-grained, so random negatives are often too easy. At the same time, continually refreshing negatives with increasingly complex policies changes the supervision distribution during training. We therefore compare a small set of negative-mining strategies under a common candidate-retrieval and evaluation protocol; strategy-specific sampling and optimization configurations are described with the corresponding experiments.

\subsection{Problem Formulation and Learning Objective}\label{sec:scoring}

Given a legal query $q$ and a document corpus $\mathcal{D}$, an embedding retriever constructs a candidate set $C_K(q) \subseteq \mathcal{D}$ containing up to $K$ documents. A reranker with parameters $\theta$ assigns a relevance score to each query--document pair and orders the candidates by decreasing score. We use $K=300$ for negative mining and standard benchmark evaluation. This controlled candidate budget allows us to study reranker adaptation after candidate retrieval; the smaller retrieval budget used for the supplementary preference evaluation is specified in Section~5.3.

For each training query, let $P(q)$ denote its known positive documents. A mining policy selects a negative set $N(q) \subseteq C_K(q) \setminus P(q)$, yielding positive and negative query--document pairs with labels $y=1$ and $y=0$, respectively. The original positives are retained for training even when they are absent from the retrieved candidate set. Excluding known positives does not guarantee that every remaining candidate is irrelevant, because relevance annotations may be incomplete. The mining strategies therefore determine which non-positive candidates provide supervision and when that supervision is refreshed.

Following the Qwen3 reranker formulation~\cite{zhang2025qwen3}, the model receives a task instruction $I$, query $q$, and document $d$, and scores relevance through the relative probabilities of the answer tokens \texttt{yes} and \texttt{no}:
\begin{equation}
 s_\theta(q,d)=
 \frac{p_\theta(\texttt{yes}\mid I,q,d)}
 {p_\theta(\texttt{yes}\mid I,q,d)+p_\theta(\texttt{no}\mid I,q,d)}.
 \label{eq:reranker_score}
\end{equation}
Writing the corresponding token logits as $\ell_\theta^{\mathrm{yes}}$ and $\ell_\theta^{\mathrm{no}}$, this score equals $\sigma(z_\theta(q,d))$, where $z_\theta(q,d)=\ell_\theta^{\mathrm{yes}}-\ell_\theta^{\mathrm{no}}$ and $\sigma$ is the sigmoid function. The score is used to order evidence by predicted relevance.

Stage1 and Stage2 optimize a pointwise binary cross-entropy objective. For a batch of $M$ labeled query--document pairs, the loss is
\begin{equation}
 \begin{aligned}
 \mathcal{L}_{\mathrm{point}}(\theta)
 =-\frac{1}{M}\sum_{i=1}^{M}\bigl[&y_i\log s_\theta(q_i,d_i)\\
 &+(1-y_i)\log\bigl(1-s_\theta(q_i,d_i)\bigr)\bigr].
 \end{aligned}
 \label{eq:pointwise_objective}
\end{equation}
This objective increases the scores of known positives and decreases those of sampled negatives. Across the two stages, the loss remains unchanged while the negative-mining model and the resulting supervision change. Our methodological focus is thus how to construct and update informative negative supervision for reranker adaptation. Section~\ref{sec:mining} defines the mining strategies, and Section~\ref{sec:designs} distinguishes independent strategy comparisons from sequential parameter inheritance.

\subsection{Hard-Negative Mining Strategies}\label{sec:mining}

Our central question is which source of hardness information provides the most effective negative supervision for reranker adaptation. All strategies draw candidate documents from embedding retrieval; they differ in the model used to identify confusing non-positive candidates and in how frequently this judgment is updated. Here, a hard negative is a candidate outside the known positive set that receives a high rank or relevance score from the mining model. Hardness is therefore relative to the model that evaluates the candidate, rather than an intrinsic property of the document.

We examine three sources of hardness information. The embedding retriever provides a natural starting point because its candidates form the input to the reranker in the retrieval pipeline. A fixed reranker offers a second source: its ranking reveals candidates that the reranker itself considers relevant despite their absence from the known positive set. This supervision may address ambiguities that embedding rankings alone do not capture. A third source is the reranker as it changes during training. As adaptation updates the model, previously difficult negatives may become easy, while other candidates remain confusing. Refreshing negatives with the evolving model may therefore keep supervision better aligned with its current ranking behavior.

For this third source, we distinguish epoch-level refresh, which holds the mined supervision fixed within a training epoch, from online refresh during training. Together with embedding-mined and fixed reranker-mined negatives, these choices yield the four strategies below. They test whether moving from retrieval-based hardness to model-specific hardness, and then to more frequently updated hardness, improves retrieval performance. Such improvements are not assumed to be monotonic: more frequent refresh also changes the supervision distribution more rapidly.

The source of hardness information is distinct from the difficulty composition of the training set. In preliminary experiments, concentrating supervision on the hardest negatives alone led to unstable training, motivating the inclusion of medium and easy negatives alongside hard ones. The purpose of this mixture is to retain challenging distinctions while also supervising less ambiguous relevance decisions. We therefore sample across difficulty regions defined by the corresponding mining policy, rather than using only the highest-ranked non-positive candidates. The following subsections describe the source and refresh mechanism of each strategy; Section~\ref{sec:designs} explains how these strategies are evaluated independently and composed sequentially.

\begin{figure}[htbp]
    \centering
    \includegraphics[width=\columnwidth,keepaspectratio]{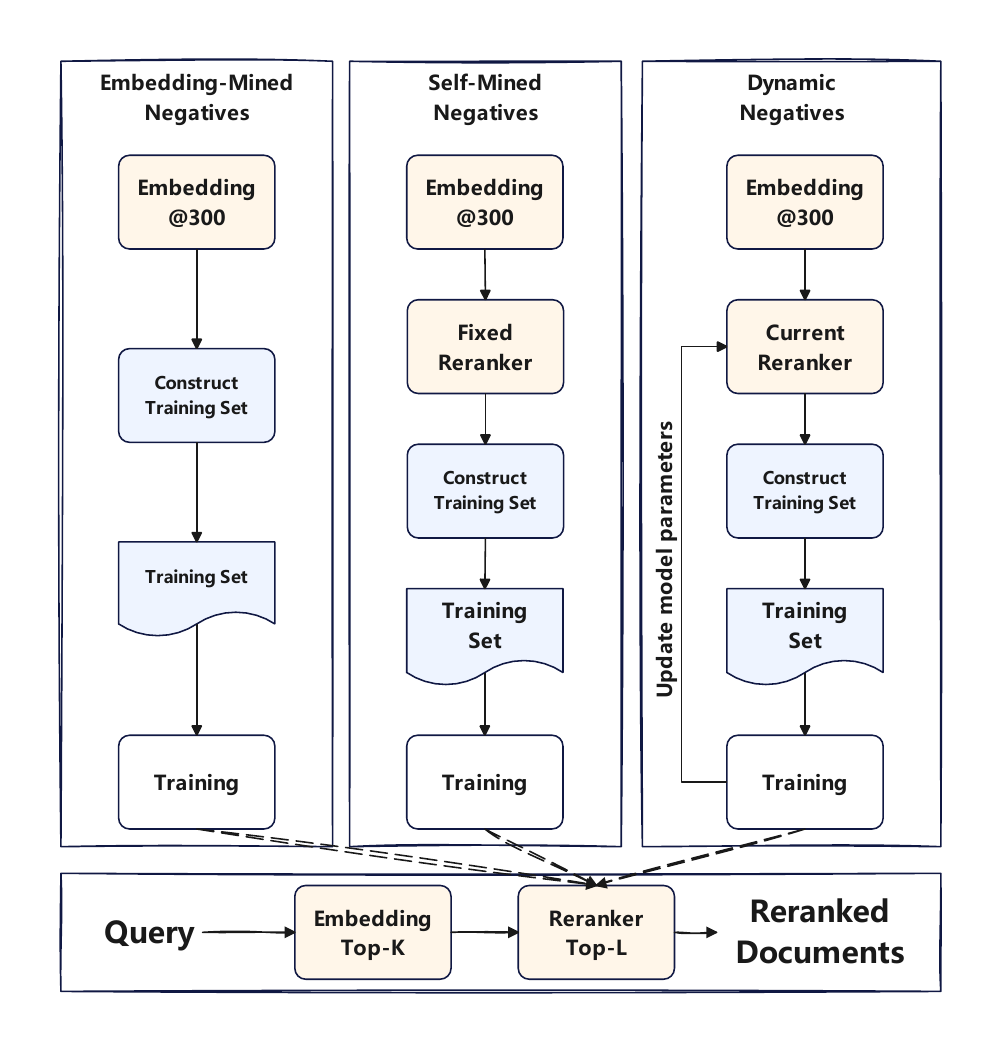}
    \caption{Overview of embedding-mined, fixed reranker-mined, and dynamic negative supervision. The columns represent alternative training strategies; the connections to the inference pipeline indicate the use of a reranker trained with one selected strategy, not a model ensemble. Training-set construction retains the original known positives, independently of candidate recall, and combines them with sampled hard, medium-difficulty, and easy negatives after excluding known positives from all negative-sampling pools. Dynamic mining includes epoch-level and online refresh, while the feedback arrow denotes model parameter updates. At inference, embedding retrieval produces top-$K$ candidates and the trained reranker returns the top-$L$ documents.}
    \label{fig:negative_mining_overview}
\end{figure}

Figure~\ref{fig:negative_mining_overview} summarizes the three sources of hardness information and their relationship to reranker training and inference.

\subsubsection{Embedding-Mined Negatives}

Embedding-mined supervision uses the first-stage retriever itself to identify difficult candidates. For each query, we retrieve the top-300 documents, exclude known positives, and remove duplicate candidates. The original embedding ranks define the difficulty regions: ranks 1--50 are treated as hard, 51--150 as medium, and 151--300 as easy. These labels describe relative difficulty within the retrieved pool. Even the easy region consists of embedding-retrieved documents rather than arbitrary documents sampled from the entire corpus.

We sample without replacement within these regions, with target proportions of approximately 50\%, 30\%, and 20\% for hard, medium, and easy negatives. For $p$ positive documents, the target negative count is $\min(\max(2p+2,6),12)$; integer quotas and candidate availability determine the realized mixture, with shortages filled from remaining eligible candidates when possible. The sampled negatives are combined with the retained positives, and the resulting supervision is fixed during the training stage. This strategy provides a natural initial adaptation step: it teaches the reranker to distinguish relevant evidence from alternatives already favored by the retriever, without requiring an adapted reranker for mining.

\subsubsection{Self-Mined Negatives}

Self-mined supervision replaces embedding rank with the judgment of a fixed reranker. The mining model scores the same embedding top-300 candidates using the relevance function in Section~\ref{sec:scoring}. After ranking, known positives and duplicates are excluded from the negative pool. A high-ranked non-positive candidate represents a distinction that this reranker finds confusing, making its ranking a potentially more task-specific source of supervision than embedding similarity. The mining model is the original base reranker in the independent comparison and the Stage1 reranker in sequential Stage2 training; the term self-mined does not imply that the mining model is updated during this stage.

The fixed reranker-mining configuration targets 12 negatives per query from reranker-rank regions 1--50, 51--150, and 151--300, using approximate hard/medium/easy proportions of 50\%/30\%/20\%. Eligible candidates are sampled without replacement, with remaining candidates used to fill shortages when available. The original positives are retained, and the mined supervision remains fixed while the target reranker is trained. This construction changes the source of the difficulty judgment while preserving a mixture of difficulty levels. It tests whether supervision aligned with a fixed reranker's confusions improves adaptation, while leaving open whether those confusions remain informative as training progresses.

\subsubsection{Epoch-Dynamic Negatives}

Epoch-dynamic mining updates the source of supervision between training rounds. At each round boundary, the current reranker scores the embedding top-300 candidate pool and constructs a new negative set after excluding known positives. The resulting examples are held fixed for the next training epoch. A selected checkpoint from that round supplies the model for the following mining pass. Thus, each round alternates an offline mining phase with a training phase, allowing the negative distribution to follow model adaptation without changing within the epoch.

Compared with fixed self-mining, this mechanism can replace negatives that have become less informative with candidates that remain confusing to the updated model. Difficulty mixing remains part of supervision construction; refreshing the model does not require training exclusively on its highest-scoring non-positive candidates. The interval between mining passes provides a period of fixed supervision, but also delays the response to changes within an epoch. This variant tests whether periodic feedback offers a useful balance between adapting the negative set and retaining a stable training distribution. Its benefit over fixed mining is assessed empirically rather than assumed from the increased refresh frequency.

\subsubsection{Online-Dynamic Negatives}

Online-dynamic mining moves negative selection into the training process. For the queries being processed, the current reranker scores the embedding top-300 candidates and selects negatives according to its current ranking and the active mining policy. These negatives are combined with positive supervision for training, so subsequent mining calls can use the updated model rather than a checkpoint fixed for an entire epoch. Mining is performed at the query-processing level; multiple query steps may be accumulated before an optimizer update. It therefore does not imply that the complete training set is re-mined after every parameter update.

This design reduces the delay between model adaptation and the identification of confusing candidates. It also couples training and supervision construction more tightly: parameter changes affect later negative selection, which in turn affects subsequent updates. The negative mixture remains important, because frequent refresh need not make an exclusively hard-negative training set well behaved. Relative to epoch-dynamic mining, the online variant provides more immediate feedback while changing the supervision distribution within an epoch. We evaluate it as a more aggressive alternative to test whether this closer tracking improves retrieval quality or instead reduces the robustness of adaptation.

\subsection{Independent and Sequential Adaptation}\label{sec:designs}

\begin{figure}[htbp]
    \centering
    \includegraphics[width=\columnwidth,keepaspectratio]{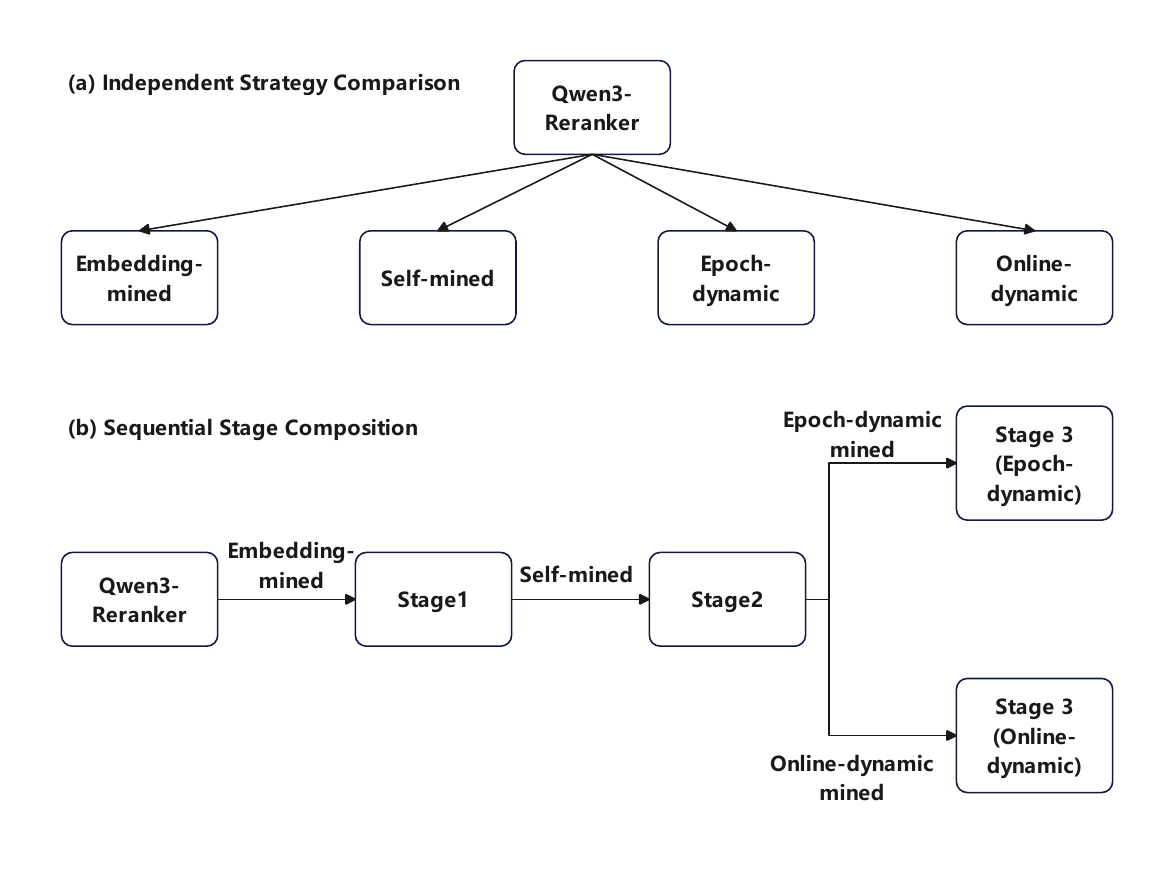}
    \caption{
    Comparison between the two experimental training designs used in this work. 
    (a) Independent strategy comparison, where each negative-mining strategy is trained separately starting from the same base reranker. 
    (b) Sequential stage composition, where Stage2 inherits Stage1 and the two Stage3 variants are alternative branches from Stage2, rather than consecutive stages.
    }
    \label{fig:training_designs}
\end{figure}

We evaluate the mining strategies under two complementary designs: independent training compares their performance from a common initialization, while sequential adaptation examines the additional benefit of changing supervision after an earlier adaptation stage. Figure~\ref{fig:training_designs} summarizes these designs, separating the choice of mining strategy from the inheritance of an adapted model.

\subsubsection{Independent Strategy Comparison}

Each independent experiment starts from the native Qwen3-Reranker-0.6B model and uses the same positive-data split, candidate budget, and standard retrieval evaluation protocol. Embedding-mined training uses retriever-ranked negatives, whereas Self-Mined training uses negatives generated by the fixed native reranker. The epoch-dynamic and online-dynamic experiments also begin from the native model, but refresh supervision using the reranker as it adapts within each experiment.

Thus, none of these experiments requires prior embedding-mined adaptation as its initialization. They compare the selected configurations of the four strategy families under a common evaluation setting. Strategy-specific sampling and optimization settings may differ, so the comparison should not be interpreted as changing only the negative source while holding every training hyperparameter fixed.

\subsubsection{Sequential Stage Composition}

The sequential design combines parameter inheritance with a change in supervision. Stage1 adapts the native reranker using embedding-mined negatives. Stage2 starts from the Stage1-adapted model and uses a new negative set mined by that model. The two Stage3 variants independently continue from Stage2 with epoch-dynamic or online-dynamic mining; neither Stage3 variant precedes the other.

At the end of each sequential stage, its LoRA weights are merged into the current reranker. The next stage starts from this merged model and trains a newly initialized LoRA adapter using its updated supervision. This design tests whether fixed self-mining adds value after embedding-based adaptation and whether subsequent dynamic refresh yields further gains. It evaluates this particular progression rather than all permutations of the mining strategies, and therefore does not establish an optimal training order. The selected two-stage recipe is subsequently evaluated at the 4B scale.

\section{Experimental Setup}

\subsection{Scaling to 4B}

After selecting the most reliable 0.6B pipeline, we scale the same recipe to a 4B reranker. The 4B model uses the same cleaned positive split and the same Stage1/Stage2 logic. Training uses LoRA adapters with rank 16 and alpha 32 on attention projection modules, a learning rate of $10^{-5}$, cosine scheduling, and maximum sequence length 1536. This scaling experiment asks whether the training policy remains useful when the base reranker is already stronger.

\subsection{Benchmarks and Metrics}

We report retrieval results on STARD~\cite{su2024stard}, CSAID~\cite{li2026legalmalrmultiagentqueryunderstandingllmbased}, and EVAL. STARD contains Chinese statute retrieval queries derived from real-life legal consultation, and CSAID reflects application-driven legal question answering and evidence retrieval; these two datasets are the final held-out benchmarks. EVAL is an internal development set for production-oriented configuration comparison and is not treated as a final benchmark.

Unless otherwise specified, standard evaluation uses embedding top-300 followed by reranking to top-30. Metrics are computed per query from this reranked list using binary known-positive labels and then averaged. MRR is the reciprocal rank of the first known positive, NDCG is computed at top-30 from discounted binary gains, and Recall@$k$ is the fraction of known positives retrieved in the first $k$ positions, for $k\in\{10,20,30\}$. This protocol isolates reranker quality after a large candidate pool has been retrieved.

The standard protocol is intentionally different from the deployment-oriented offline preference setting described below. In standard evaluation, top-300 retrieval is used to reduce the chance that reranker comparisons are dominated by missing candidates. In preference evaluation, top-100 to top-5 is used because it better reflects a latency-sensitive product configuration where users or downstream generators consume only a compact evidence list.

\subsection{Operational-Query Preference Evaluation}

Benchmark metrics may not fully reflect whether the returned evidence list is useful in a deployment-oriented retrieval setting. We therefore conduct a supplementary offline preference evaluation on 410 de-identified and reconstructed internal operational-query inputs after deduplication and basic noise filtering. The inputs contain no names, account identifiers, contact information, device identifiers, timestamps, precise locations, or other information retained for identifying users. We do not analyze user clicks, individual behavioral trajectories, or user-level feedback.

For each query, we compare the native 4B reranker and the trained 4B Stage2 reranker under a latency-sensitive online-style retrieval setting: embedding top-100 followed by reranking to top-5. This configuration is intentionally smaller than the standard top-300 evaluation protocol because practical search and RAG systems typically expose or consume only a compact evidence list.

The two systems are anonymized as A and B with randomized order using a fixed seed. We use Qwen3.6-Max as the LLM judge. Under a fixed judging prompt, it receives the query and both top-5 evidence lists and returns exactly one label, A, B, or Tie, according to relevance, legal specificity, coverage, and lack of distracting evidence. The judging protocol and a subset of LLM-labeled comparisons were reviewed and spot-checked by internal researchers for quality control; these manual checks were not used as the primary evaluation outcome. We report only aggregate comparison statistics and use a two-sided exact binomial sign test under the null hypothesis that the trained model wins half of non-tied comparisons.

\section{Results}

\subsection{Negative-Mining Strategy Comparison}

Table~\ref{tab:ablation} compares the four mining strategies when trained independently from the native 0.6B reranker, as described in Section~\ref{sec:designs}. The comparison examines whether useful negative supervision is better identified by embedding ranks, a fixed reranker, or an evolving reranker. All models are evaluated under the same candidate budget and ranking protocol. Because sampling and optimization settings also differ between strategy configurations, the results characterize these configurations rather than isolate the causal effect of the negative source alone. The embedding-mined model in this table and Stage1 in Table~\ref{tab:stage_progression} use the same Stage1 checkpoint, but their results were obtained in separate retrieval and inference runs; the small numerical differences between the two tables should therefore not be interpreted as different trained models.

\begin{table*}[t]
\caption{0.6B negative-mining ablation under the aligned standard evaluation protocol.}
\centering
\begin{tabular}{llrrrrr}
\toprule
Dataset & Model & MRR & NDCG & R@10 & R@20 & R@30 \\
\midrule
EVAL & Base Reranker & 0.6651 & 0.6814 & 0.7839 & 0.8490 & 0.8717 \\
 & Embedding-mined & 0.7056 & 0.7079 & 0.7829 & 0.8491 & 0.8630 \\
 & Self-Mined  & \textbf{0.7081} & 0.7083 & \textbf{0.7882} & 0.8413 & 0.8595 \\
 & Epoch-dynamic & 0.7010 & \textbf{0.7146} & 0.7826 & \textbf{0.8557} & \textbf{0.8818} \\
 & Online-dynamic & 0.6897 & 0.6906 & 0.7609 & 0.8056 & 0.8192 \\
\midrule
STARD & Base Reranker & 0.5778 & 0.5883 & 0.6651 & 0.7543 & 0.7884 \\
 & Embedding-mined & 0.5568 & 0.5284 & 0.6149 & 0.6304 & 0.6375 \\
 & Self-Mined  & 0.5611 & 0.5228 & 0.5970 & 0.6079 & 0.6153 \\
 & Epoch-dynamic & \textbf{0.6172} & \textbf{0.6111} & \textbf{0.6887} & \textbf{0.7546} & \textbf{0.7895} \\
 & Online-dynamic & 0.6030 & 0.5698 & 0.6351 & 0.6510 & 0.6684 \\
\midrule
CSAID & Base Reranker & 0.6635 & 0.5168 & 0.4097 & 0.5531 & 0.6270 \\
 & Embedding-mined & 0.8013 & \textbf{0.5603} & \textbf{0.4278} & \textbf{0.5610} & \textbf{0.6436} \\
 & Self-Mined  & 0.7941 & 0.5543 & 0.4274 & 0.5487 & 0.6190 \\
 & Epoch-dynamic & 0.7781 & 0.5447 & 0.4211 & 0.5502 & 0.6210 \\
 & Online-dynamic & \textbf{0.8016} & 0.5076 & 0.3958 & 0.4906 & 0.5221 \\
\bottomrule
\end{tabular}
\label{tab:ablation}
\end{table*}

Fixed mining provides substantial gains on CSAID. Embedding-mined training raises MRR from 0.6635 to 0.8013 and achieves the highest NDCG and recall at all three cutoffs among the compared configurations. Self-Mined training also improves CSAID MRR, to 0.7941, but does not surpass embedding-mined training on this dataset. On EVAL, Self-Mined training yields the highest MRR and Recall@10, although its MRR advantage over embedding-mined training is small (0.7081 versus 0.7056). These results establish embedding-ranked candidates as an effective source of negative supervision, while showing that replacing their difficulty ordering with a fixed reranker's ordering does not yield a uniform benefit.

The dynamic strategies differ substantially. Epoch-dynamic mining achieves the highest values for all reported STARD metrics and for EVAL NDCG, Recall@20, and Recall@30. In contrast, both fixed-mining strategies reduce every reported STARD metric relative to the native model. Online-dynamic mining attains the highest CSAID MRR, 0.8016, but lowers NDCG and all three recall measures below the native baseline; Recall@30 falls from 0.6270 to 0.5221. A high MRR can therefore coexist with poorer retrieval of the remaining relevant documents: it rewards the position of the first relevant result, whereas recall reflects how much relevant material is retained within the cutoff. The online variant's MRR alone would give an incomplete account of its evidence-selection quality.

No configuration dominates across datasets and metrics. Periodic refresh is effective on STARD, whereas embedding-mined supervision remains competitive on CSAID. More frequent refresh does not consistently improve either ranking quality or recall. These observations motivate examining whether fixed and dynamic supervision can add value after an initial adaptation stage, rather than assuming that the strongest independent strategy will also be the best subsequent training stage.

\subsection{Sequential Hard-Negative Adaptation}

Table~\ref{tab:stage_progression} evaluates the progression defined in Section~\ref{sec:designs}: embedding-mined adaptation in Stage1, fixed self-mining with the Stage1 model in Stage2, and two alternative dynamic-mining branches from Stage2. Each successive stage inherits the preceding stage's merged model and trains a new LoRA adapter. This experiment assesses the incremental value of changing supervision along the specified progression; it does not compare all possible training orders.

\begin{table*}[t]
\caption{Sequential stage progression on the Qwen3-0.6B reranker backbone. Stage2 provides a strong trade-off, while additional Stage3 variants are not uniformly better.}
\centering
\small
\begin{tabular}{llrrrrr}
\toprule
Dataset & Model & MRR & NDCG & R@10 & R@20 & R@30 \\
\midrule
EVAL & Base Reranker & 0.6651 & 0.6814 & 0.7839 & \textbf{0.8490} & \textbf{0.8717} \\
 & Stage1 & 0.7066 & 0.7085 & 0.7839 & 0.8472 & 0.8630 \\
 & Stage2 & 0.7124 & 0.7132 & \textbf{0.7918} & 0.8444 & 0.8597 \\
 & Stage3 (epoch-dynamic) & \textbf{0.7270} & \textbf{0.7199} & 0.7911 & 0.8356 & 0.8463 \\
 & Stage3 (online dynamic) & 0.7238 & 0.7188 & 0.7900 & 0.8346 & 0.8460 \\
\midrule
STARD & Base Reranker & \textbf{0.5778} & \textbf{0.5883} & \textbf{0.6651} & \textbf{0.7543} & \textbf{0.7884} \\
 & Stage1 & 0.5570 & 0.5288 & 0.6149 & 0.6304 & 0.6375 \\
 & Stage2 & 0.5310 & 0.4910 & 0.5494 & 0.5603 & 0.5632 \\
 & Stage3 (epoch-dynamic) & 0.5162 & 0.4734 & 0.5231 & 0.5286 & 0.5305 \\
 & Stage3 (online dynamic) & 0.5093 & 0.4676 & 0.5158 & 0.5202 & 0.5221 \\
\midrule
CSAID & Base Reranker & 0.6635 & 0.5168 & 0.4097 & 0.5531 & 0.6270 \\
 & Stage1 & 0.7971 & 0.5600 & 0.4278 & 0.5624 & 0.6445 \\
 & Stage2 & 0.8156 & \textbf{0.5894} & \textbf{0.4672} & \textbf{0.6042} & \textbf{0.6641} \\
 & Stage3 (epoch-dynamic) & 0.8260 & 0.5771 & 0.4610 & 0.5834 & 0.6337 \\
 & Stage3 (online dynamic) & \textbf{0.8265} & 0.5776 & 0.4598 & 0.5757 & 0.6356 \\
\bottomrule
\end{tabular}
\label{tab:stage_progression}
\end{table*}

Moving from Stage1 to Stage2 improves all reported CSAID metrics: MRR increases from 0.7971 to 0.8156, NDCG from 0.5600 to 0.5894, and Recall@10 from 0.4278 to 0.4672. On EVAL, MRR, NDCG, and Recall@10 also improve, but Recall@20 and Recall@30 decline slightly. Thus, the additional stage benefits early ranking on EVAL and both ranking and coverage within the evaluated cutoffs on CSAID. This pattern is consistent with self-mining supplying useful supervision after embedding-based adaptation. However, the stage comparison also includes additional training, so it does not attribute the gains exclusively to the change in negative source.

Stage3 introduces a clearer trade-off between early relevance and broader evidence retention. The epoch-dynamic branch increases EVAL MRR from 0.7124 to 0.7270 and NDCG from 0.7132 to 0.7199, while reducing recall at every reported cutoff relative to Stage2. Both Stage3 branches likewise raise CSAID MRR but reduce its NDCG and all recall measures. Stage2 retains the highest CSAID NDCG and recall, as well as the highest EVAL Recall@10, within the progression. The additional dynamic stages therefore improve some aspects of ranking without consistently improving the evidence list as a whole.

STARD exposes a further limitation: every adaptation stage performs below the native model, and performance declines along the progression. MRR falls from 0.5570 at Stage1 to 0.5310 at Stage2, then to 0.5162 and 0.5093 in the epoch-dynamic and online-dynamic branches, respectively. Notably, epoch-dynamic mining performs well when trained independently in Table~\ref{tab:ablation}, but does not recover STARD performance when applied after Stage2. Its utility therefore depends on the preceding adaptation as well as the mining configuration. The pattern is consistent with increasing specialization to the training distribution, although these endpoint evaluations do not establish the underlying mechanism.

We select Stage2 for the subsequent scaling experiments because it provides the highest EVAL Recall@10 among the staged configurations and a practical development-stage trade-off between early ranking and evidence retention. This is not a claim of uniform superiority across benchmarks: the native reranker remains strongest on STARD, and Stage3 achieves higher MRR on EVAL and CSAID. STARD and CSAID remain the final held-out benchmark evaluations, whereas EVAL is used only for development-stage configuration comparison. The results support retaining fixed self-mining after embedding-based adaptation in the selected pipeline, while treating further dynamic refresh as an empirical choice whose benefit depends on the target dataset and ranking metric.

\subsection{Scaling the Pipeline to 4B}

Table~\ref{tab:scaling} shows that the selected training recipe scales to a stronger base model. The native 4B reranker is already stronger than the 0.6B Stage2 model, but applying the two-stage training pipeline still improves the 4B model across datasets. On EVAL, MRR increases from 0.7356 to 0.7982; on CSAID, MRR increases from 0.8369 to 0.9094 and Recall@10 from 0.5735 to 0.6625. These gains suggest that hard-negative reranker training remains valuable even when starting from a stronger general-purpose reranker.

\begin{table*}[t]
\caption{Comparison of the 0.6B Stage2 model, native 4B reranker, and trained 4B Stage2 model under the aligned standard evaluation protocol.}
\centering
\small
\begin{tabular}{llrrrrr}
\toprule
Dataset & Model & MRR & NDCG & R@10 & R@20 & R@30 \\
\midrule
EVAL & 0.6B Stage2 & 0.7124 & 0.7132 & 0.7918 & 0.8444 & 0.8597 \\
 & 4B Base & 0.7356 & 0.7616 & 0.8749 & 0.9301 & 0.9477 \\
 & 4B Stage2 & \textbf{0.7982} & \textbf{0.8147} & \textbf{0.9259} & \textbf{0.9542} & \textbf{0.9612} \\
\midrule
STARD & 0.6B Stage2 & 0.5310 & 0.4910 & 0.5494 & 0.5603 & 0.5632 \\
 & 4B Base & 0.6783 & 0.6851 & 0.7736 & 0.8384 & \textbf{0.8816} \\
 & 4B Stage2 & \textbf{0.7126} & \textbf{0.7062} & \textbf{0.7918} & \textbf{0.8579} & \textbf{0.8816} \\
\midrule
CSAID & 0.6B Stage2 & 0.8156 & 0.5894 & 0.4672 & 0.6042 & 0.6641 \\
 & 4B Base & 0.8369 & 0.6942 & 0.5735 & 0.7392 & 0.8160 \\
 & 4B Stage2 & \textbf{0.9094} & \textbf{0.7627} & \textbf{0.6625} & \textbf{0.8095} & \textbf{0.8514} \\
\bottomrule
\end{tabular}
\label{tab:scaling}
\end{table*}

\subsection{Operational-Query Preference Results}

Table~\ref{tab:ab_test} reports the operational-query preference evaluation under the Qwen3.6-Max LLM-as-judge protocol described above. The trained 4B Stage2 reranker wins 239 comparisons, loses 91, and ties 80. Among the 330 non-tied comparisons, its win rate is 72.4\%, with a 95\% Wilson confidence interval of 67.4--77.0\%. The two-sided exact sign test rejects equal win probability among non-tied comparisons ($p = 3.11 \times 10^{-15}$).

\begin{table*}[t]
\caption{LLM-based pairwise preference evaluation on 410 de-identified, reconstructed internal operational-query inputs. Win rate and its 95\% Wilson confidence interval (CI) exclude ties; the $p$-value is from a two-sided exact sign test.}
\centering
\small
\begin{tabular}{lrrrrrr}
\toprule
Comparison & $n$ & Win & Tie & Lose & Win Rate (95\% CI) & $p$ \\
\midrule
4B Stage2 vs 4B Base & 410 & 239 & 80 & 91 & 72.4\% [67.4\%, 77.0\%] & $3.11 \times 10^{-15}$ \\
\bottomrule
\end{tabular}
\label{tab:ab_test}
\end{table*}

Ties account for 19.5\% of all comparisons. When they are retained in the denominator, Stage2 wins 58.3\% of the 410 comparisons; the reported 72.4\% win rate is conditional on the judge expressing a preference. The confidence interval describes uncertainty in this conditional win proportion and does not account for systematic judge bias.

This evaluation uses embedding top-100 followed by reranking to top-5, reflecting the compact evidence lists consumed in latency-sensitive legal search and RAG systems. The preference for Stage2 provides supplementary support for its evidence-selection quality under this smaller retrieval budget. The measured outcome is the LLM judge's preference between evidence lists under the specified criteria; generated-answer quality and user satisfaction are not directly evaluated.

\section{Discussion}

The main lesson is that industrial retrieval optimization should be component-aware. In our setting, embedding recall under a large candidate pool shows diminishing gains, while reranking still determines which retrieved evidence reaches the final list. Reranker supervision is therefore a useful target for separate evaluation under a fixed candidate budget. Dynamic negative mining provides informative ablations, but the evaluated dynamic stages do not improve all metrics relative to the selected two-stage configuration.

Our evaluation is intentionally scoped to Chinese legal retrieval and includes supplementary de-identified, reconstructed operational-query inputs from the deployed application setting. We do not claim that the same numerical conclusions transfer unchanged to general-domain retrieval. Instead, the broader implication is methodological: before adding complexity, practitioners should measure whether candidate recall is still the bottleneck, then choose negative-mining policies according to aligned held-out evaluation and operational-query evidence quality.

The STARD results also differ between the 0.6B and 4B settings. In the 0.6B progression, in-domain hard-negative tuning improves CSAID and some EVAL metrics but lowers STARD metrics. At 4B, the selected Stage2 model improves STARD over the native 4B reranker while also improving CSAID and EVAL. This contrast indicates that the observed transfer pattern depends on the model scale and the evaluated configuration; the present experiments do not identify the mechanism responsible for the difference.

More broadly, our results suggest that retrieval optimization may be phase-dependent rather than uniformly monotonic. When candidate recall is still limited, expanding embedding coverage can remain beneficial. When recall gains flatten under a sufficiently large candidate pool, reranker supervision warrants separate evaluation as a complementary optimization target. Beyond this point, increasingly aggressive hard-negative adaptation may improve some metrics while reducing others across query distributions.

There are also limitations. The internal validation split is a random sample from the cleaned training pool and is used only for early stopping, not as the primary test set. We remove exact query overlaps with held-out benchmark queries before negative mining, but we do not claim case-level, document-level, or temporal isolation. In addition, the operational-query preference evaluation uses an LLM judge with internal spot-checking. We therefore treat it as deployment-oriented supplementary evidence, not a replacement for controlled benchmark metrics.

\section{Conclusion}

We studied a two-stage industrial legal retrieval pipeline under a large candidate pool with high embedding recall. Across embedding analysis, negative-mining ablations, 0.6B-to-4B scaling, and operational-query preference evaluation, the results show that reranker adaptation merits separate evaluation when recall gains flatten. Among the evaluated staged configurations, the two-stage hard-negative pipeline provides a practical trade-off, while additional dynamic stages improve some metrics but not others. In particular, adaptive hard-negative refresh is not a monotonic improvement for reranker training under the high-recall candidate-pool setting evaluated here. The trained 4B Stage2 reranker improves the final STARD and CSAID benchmark metrics and is preferred in the supplementary internal operational-query evaluation, supporting its use as one practical configuration for legal retrieval.

The broader implication is that industrial retrieval systems should assess candidate coverage and final evidence ordering separately, rather than assuming that continued recall-oriented scaling or additional adaptation stages will consistently improve downstream retrieval quality.

\bibliographystyle{IEEEtran}
\bibliography{references}
\end{document}